\documentclass[letterpaper, 10pt, conference]{IEEEconf}  
\usepackage[utf8]{inputenc}

\usepackage{xcolor}  
\usepackage{graphicx} 

\usepackage{amsmath} 
\usepackage{amssymb}  
\usepackage{wasysym}
\usepackage{dsfont}     
\usepackage{siunitx}
\usepackage{graphics}
\usepackage{algorithm}  
\usepackage{algpseudocode}  

\makeatletter
\let\NAT@parse\undefined
\makeatother
\usepackage[pdftex,
        colorlinks=true,
        bookmarks=true,
        citecolor=red,
        linkcolor=blue,
        pagebackref=false,
        pdfstartview=Fit,
        breaklinks=true
            ]{hyperref}
\usepackage{cite}
\usepackage{cleveref}
\usepackage{lineno}
\usepackage{tabularx}
\usepackage{booktabs}
\usepackage{setspace}
\usepackage{threeparttable}
\usepackage{tikz}
\usepackage{multirow}
\usepackage{cuted}

\usepackage{caption}
\usepackage{stfloats}
\usepackage{float}

\IEEEoverridecommandlockouts                              

\title{\LARGE \bf

RoboCasa-Shift: Stage-Aware Robustness Evaluation of Vision-Language-Action Policies in Long-Horizon Household Manipulation
}

\author{Haocheng~Wang\textsuperscript{1,3},
Youquan~Liu\textsuperscript{1},
Weiyu~Guo\textsuperscript{2},
Aokai~Zhang\textsuperscript{3},
Donghang~Wu\textsuperscript{4},
Hongyu~Liu\textsuperscript{5},\\
Sha Zhang\textsuperscript{2},
Weidong~Yang\textsuperscript{1\cross},
Ben~Fei\textsuperscript{2\cross}
\thanks{
\textsuperscript{1} Fudan University.
\textsuperscript{2} The Chinese University of Hong Kong.
\textsuperscript{3} Shenzhen Loop Area Institute, Shenzhen, China. 
\textsuperscript{4} University of Chinese Academy of Sciences.
\textsuperscript{5} Shanghai Jiao Tong University.
}
}

\makeatletter

\newcommand{\Rmnum}[1]{\expandafter\@slowromancap\romannumeral #1@}

\newcommand{\cross}{\textsuperscript{\dag}}
\begin{document}

\maketitle

\thispagestyle{empty}
\pagestyle{empty}

\begin{abstract}
Vision-language-action (VLA) robustness is commonly summarized by aggregate task success, which obscures how individual executions change under distribution shifts and where failures emerge in long-horizon tasks. We introduce \textbf{RoboCasa-Shift}, a diagnostic robustness benchmark built on RoboCasa365 that combines controlled single-factor shifts, matched clean-perturbed rollouts in which each pair shares the same underlying settings and differs only in the target shift, and simulator-grounded stage-aware diagnostics. We evaluate GR00T N1.5 on four long-horizon tasks that pass a predefined clean-performance gate under four perturbation factors and three calibrated severity levels, including conditions that drive the evaluated policy to complete task failure, totaling 1,040 formal rollouts. Robustness responses vary substantially across factors and tasks, and increasing severity does not always produce monotonic degradation. For example, one camera condition yields $\Delta\mathrm{SR}=0$ despite 8 of 80 matched episodes changing outcome. Stage-aware analysis further shows that the same end-to-end degradation can correspond to different execution-depth profiles. Camera $L_2$ and Gaussian blur $L_1$ both yield $\Delta\mathrm{SR}=0.188$, yet their first-stage failure rates are 19.0\% and 36.4\%, respectively. Our findings suggest that VLA robustness should be viewed as a multi-dimensional property of execution rather than a single end-to-end performance measure, motivating evaluation protocols that jointly capture outcome degradation, paired outcome sensitivity, and failure progression.
The code is publicly available at \url{https://github.com/HaochengWang23/RoboCasa-Shift}.
\end{abstract}


\section{Introduction}

\begin{figure*}[htbp]
    \vspace{0.5em}
    \centering
    \includegraphics[width=\textwidth]{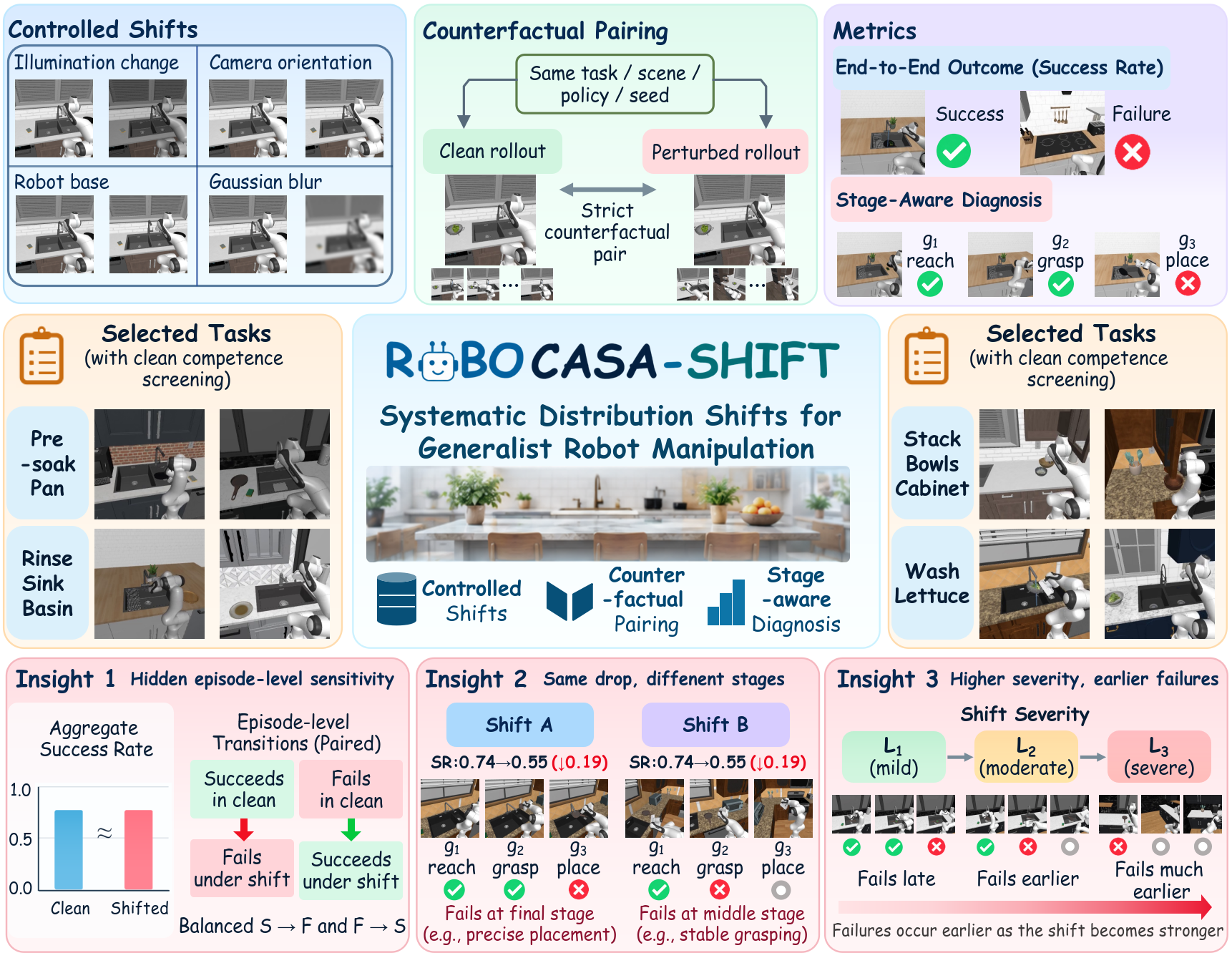}
    \caption{\textbf{Overview of RoboCasa-Shift.}
    RoboCasa-Shift evaluates VLA robustness on selected long-horizon RoboCasa tasks under controlled single-factor distribution shifts with multiple frozen severity levels.
    Strictly matched clean-perturbed rollouts enable counterfactual comparison under shared task conditions, while robustness is characterized jointly through end-to-end task outcomes and stage-aware semantic execution diagnostics. 
    Representative findings illustrate that aggregate success rates can mask substantial episode-level outcome changes, similar end-to-end degradation can correspond to different failure locations, and severe visual corruption can shift failures toward earlier stages of task execution.}
    \label{fig:overview}
    \vspace{-1.0em}
\end{figure*}

Vision-language-action (VLA) models have become an increasingly important paradigm for robotic manipulation by integrating visual observations, language instructions, and robot actions within unified policy frameworks \cite{zitkovich2023rt,kim2024openvla}. Despite their growing capabilities, maintaining reliable behavior under distribution shifts remains a key challenge for deployment, as robots may encounter variations in observations, environments, and physical states that differ from those seen during training.
Systematically evaluating VLA robustness under such distribution shifts is important for understanding the reliability of these models in robotic applications.


Recent robustness benchmarks have expanded the evaluation of VLA policies beyond nominal task settings through diverse controlled distribution shifts and increasingly systematic evaluation protocols \cite{robocasa365,fei2025libero,wang2026libero}.
Despite this broader coverage, robustness is still commonly summarized at the task or evaluation-condition level through aggregate metrics such as success rate. Such summaries do not reveal how individual executions change under a distribution shift or where failures emerge within long-horizon task execution, leaving a need for more diagnostic evaluation at the execution level.

Such diagnosis requires controlling episode-level variation for meaningful comparison, since clean and shifted evaluations with different task instances or initial conditions may confound episode difficulty with the target distribution shift.
We therefore compare each perturbed execution with a corresponding clean execution under the same underlying task conditions, changing only the target perturbation. This counterfactual pairing provides a controlled basis for identifying episode-level outcome changes associated with the evaluated shift.

Even under such controlled comparisons, a binary task outcome provides only a coarse description of failure in long-horizon manipulation. These tasks unfold through a sequence of semantically meaningful subgoals~\cite{dai2026robomme,han2026robocerebra}, so two failed executions may terminate after very different amounts of task progress.
Stage-aware evaluation can therefore complement final success or failure by tracking semantic progress and identifying the first incomplete subgoal, providing a clearer view of where robustness failures emerge during execution.

We introduce RoboCasa-Shift, a diagnostic robustness benchmark built on RoboCasa365~\cite{robocasa365} for evaluating VLA policies under controlled distribution shifts in long-horizon household manipulation, as illustrated in Figure~\ref{fig:overview}.
The benchmark defines controlled single-factor shifts at multiple frozen severity levels and a matched clean-perturbed protocol for episode-level counterfactual comparison.
It further incorporates simulator-grounded stage-aware diagnostics to track semantic execution progress and localize where failures emerge during long-horizon task execution. RoboCasa-Shift characterizes robustness from complementary perspectives, including end-to-end task outcomes, episode-level outcome changes, and stage-aware execution behavior.

Our evaluation reveals that robustness responses vary across perturbation factors and tasks, and that increasing calibrated severity does not always lead to monotonic degradation in end-to-end performance.
Aggregate success rates can also mask considerable episode-level outcome changes, while similar end-to-end degradation can correspond to failures emerging at different semantic depths of task execution. 
Stage-aware analysis further shows that perturbation severity can alter the semantic depth reached by failed executions, with stronger shifts in some conditions associated with earlier incomplete milestones.

The main contributions of this work are as follows:

\begin{itemize}
    \item We develop \textbf{RoboCasa-Shift}, a diagnostic robustness benchmark built on RoboCasa365 for long-horizon household 
    manipulation. The benchmark defines controlled single-factor distribution shifts at multiple frozen severity levels under an
    episode-level clean-perturbed counterfactual protocol, enabling robustness changes to be examined under matched task conditions.

    \item We introduce a simulator-grounded stage-aware robustness diagnostic that complements final task success with semantic execution information. Long-horizon tasks are represented through ordered semantic
    subgoals evaluated by simulator-state predicates, allowing execution progress and the first incomplete subgoal to be tracked for failure localization.

    \item We conduct a controlled robustness study on four clean-qualified long-horizon RoboCasa tasks across four perturbation factors and three severity levels, jointly analyzing end-to-end task outcomes, counterfactual episode-level changes, and stage-aware execution depth. The analysis reveals
    robustness behaviors not captured by aggregate success alone, including heterogeneous severity responses and differences in 
    the semantic depth reached by failed executions.
\end{itemize}
\section{Related Work}
\subsection{Vision-Language-Action Models for Robot Manipulation}
\label{sec:related_vla}

VLA models have emerged as a promising paradigm for robot manipulation by integrating visual perception, language understanding,
and action prediction within a unified policy framework
\cite{zitkovich2023rt,kim2024openvla}. 
Building on large-scale and heterogeneous robot datasets, subsequent works have
further developed generalist policies that can operate across diverse tasks, robot embodiments, and observation or action spaces
\cite{team2024octo,li2026matters}. 
Recent studies have continued to extend VLA models toward more complex sequential manipulation by improving temporal reasoning and visual grounding during task execution
\cite{xiao2026ava,liu2026activevla}. 
As VLA policies expand to diverse and complex manipulation settings,  evaluating their reliability under distribution shifts becomes increasingly important.

\subsection{Robustness and Generalization Benchmarks for Robot Manipulation}
\label{sec:related_robustness}

Existing robot manipulation benchmarks have increasingly emphasized generalization beyond the training distribution by evaluating policies across diverse tasks, environments, and controlled variations
\cite{liu2023libero,li2024evaluating}.
More recent benchmarks have substantially expanded this direction by evaluating generalist robot policies under increasingly diverse and systematically controlled distribution shifts
\cite{robocasa365,fei2025libero,wang2026libero}.
These benchmarks support systematic analysis across perturbation factors, difficulty levels, and task variations, providing broad measurements of policy robustness and generalization. However, such analyses predominantly characterize performance
at the task or evaluation-condition level, providing limited information about how distribution shifts affect the semantic execution process within a long-horizon task. RoboCasa-Shift complements these efforts through controlled distribution-shift evaluation, episode-level counterfactual comparison, and semantic subgoal-aware diagnosis of robustness degradation in long-horizon execution.


\subsection{Long-Horizon and Compositional Manipulation Evaluation}
\label{sec:related_long_horizon}
Recent progress in robot manipulation evaluation has extended benchmark tasks from short-horizon skills to long-horizon and multi-stage executions that require policies to coordinate multiple interactions over time
\cite{li2023behavior,nasiriany2024robocasa,11444456}.
Beyond increasing task horizon, several benchmarks further expose the temporal or subtask structure of manipulation tasks to evaluate capabilities such as planning, sequential reasoning, and memory-dependent execution
\cite{han2026robocerebra,dai2026robomme}. 
Despite these advances, such task structure is primarily used to characterize long-horizon manipulation capabilities themselves, while its role as a diagnostic unit for robustness under controlled distribution shifts remains less explored. RoboCasa-Shift builds on this structured view of long-horizon execution by using semantic subgoals to track execution progress and characterize the
semantic depth under distribution shifts.


\section{RoboCasa-Shift Benchmark}



\begin{figure}[t]
    \vspace{0.5em}
    \centering
    \includegraphics[width=\columnwidth]{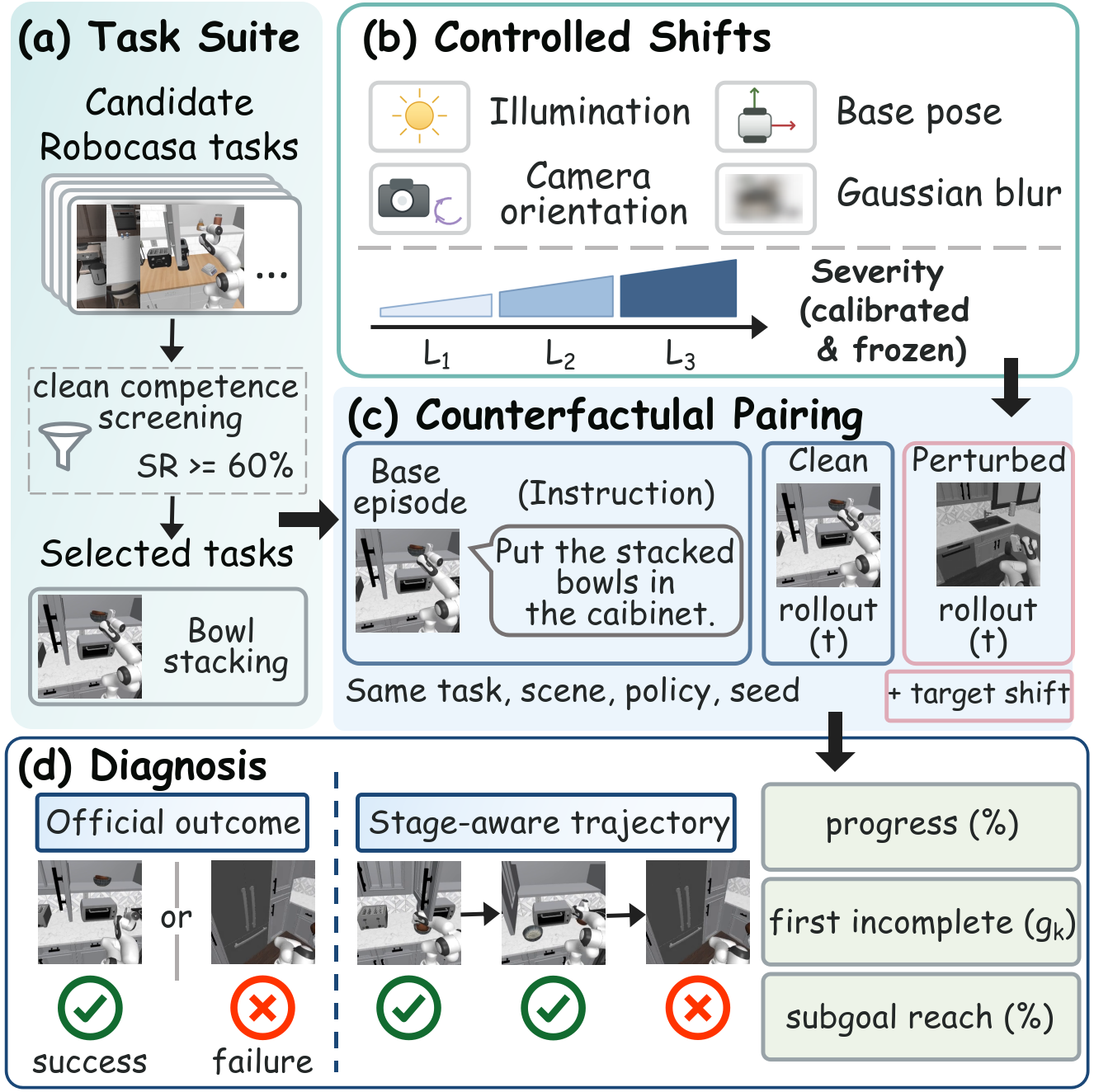}
    \caption{\textbf{Pipeline of RoboCasa-Shift.}
    (a) Task Suite. Candidate RoboCasa tasks are filtered through clean-competence screening to construct the formal evaluation suite.
    (b) Controlled Shifts. Selected tasks are evaluated under four single-factor distribution shifts with calibrated and frozen severity levels.
    (c) Counterfactual Pairing. Clean and perturbed rollouts share the underlying task conditions and policy seed and differ only in the target shift.
    (d) Diagnosis. RoboCasa's official task outcome is complemented by stage-aware semantic diagnostics of execution progress, first incomplete subgoal, and subgoal reach.}
    \label{fig:method_overview}
    \vspace{-1.5em}
\end{figure}

Figure~\ref{fig:method_overview} summarizes the overall evaluation pipeline.
RoboCasa-Shift defines a diagnostic robustness evaluation protocol for long-horizon household manipulation. It combines clean-competence task screening, controlled single-factor shifts with frozen severity levels, and matched clean-perturbed counterfactual rollouts. RoboCasa's official task-success criterion is retained for end-to-end evaluation, while semantic subgoals tracked through simulator-state predicates provide complementary stage-aware diagnostics of execution progress. The formal study in this paper
instantiates this protocol on the clean-qualified task suite described below.

\begin{table}[t]
    \vspace{0.5em}
    \centering
    \caption{\textbf{Formal task suite used in RoboCasa-Shift.}
    The official horizon, mobility label, and task-structure value are obtained
    from RoboCasa365 task metadata. Diagnostic milestones denote the
    task-specific semantic execution milestones used for stage-aware analysis
    and are not used as task-complexity labels.}
    \label{tab:task_suite}

    \small
    \renewcommand{\arraystretch}{1.05}
    \setlength{\tabcolsep}{3pt}

    \begin{tabularx}{\columnwidth}{
        @{}
        >{\raggedright\arraybackslash}p{0.27\columnwidth}
        >{\centering\arraybackslash}p{0.11\columnwidth}
        >{\centering\arraybackslash}p{0.21\columnwidth}
        >{\centering\arraybackslash}p{0.15\columnwidth}
        >{\centering\arraybackslash}p{0.17\columnwidth}
        @{}
    }
        \toprule
        \textbf{Task} &
        \textbf{Horizon} &
        \textbf{Mobility} &
        \textbf{Official Structure} &
        \textbf{Diagnostic Milestones} \\
        \midrule
        PreSoakPan        & 2400 & Non-mobile & 3 & 3 \\
        RinseSinkBasin    & 1350 & Non-mobile & 2 & 3 \\
        StackBowlsCabinet & 2100 & Mobile     & 2 & 2 \\
        WashLettuce       & 1650 & Non-mobile & 2 & 3 \\
        \bottomrule
    \end{tabularx}
    \vspace{-1.5em}
\end{table}

\subsection{Task Suite}
\label{sec:task_suite}

We screen 16 pre-frozen RoboCasa365 Composite-Seen tasks
\cite{robocasa365}, with ordering and clean-performance
thresholds fixed before any perturbation results are observed.
To avoid conflating perturbation-induced
failures with insufficient nominal task competence, we apply a predefined
clean-performance screen before perturbation evaluation. Each candidate is
first evaluated over 10 clean rollouts; tasks with at most three successes are
excluded as clean-floor cases, while the remaining candidates are extended to
20 rollouts. We retain all and only tasks achieving at least 12/20 clean successes, resulting in the four-task formal suite summarized in Table~\ref{tab:task_suite}. 
The complete screening results are provided in Appendix~\ref{app:task_selection}.
The reported diagnostic milestone count is used only for stage-aware analysis and is distinct from official task-structure metadata, as described in Section~\ref{sec:stage_tracking}.

\subsection{Controlled Distribution Shifts}
\label{sec:controlled_shifts}

RoboCasa-Shift considers four representative single-factor shifts that preserve the task objective while perturbing distinct sources of variation in household manipulation. Following robustness dimensions explored in recent VLA manipulation benchmarks~\cite{robocasa365,fei2025libero,zhang2026world}, we evaluate illumination, third-person camera orientation, robot base initialization, and Gaussian blur. For $L_1$--$L_3$, illumination scales simulator light intensity by $\alpha=\{0.8,0.6,0.4\}$; camera orientation samples third-person yaw/pitch offsets within $\theta_{\max}=\{2^\circ,6^\circ,10^\circ\}$ while preserving the wrist view; base initialization uses
$\sigma_{xy}=\{0.02,0.05,0.10\}$\,m and
$\sigma_{\psi}=\{2^\circ,5^\circ,10^\circ\}$ for planar position and yaw; and Gaussian blur applies $\sigma=\{1,2,4\}$ to the RGB observations used by the policy.
Implementation details and severity calibration procedures are provided in
Appendix~\ref{app:perturbation_details}.


Severity levels are calibrated before formal evaluation through model-free rendering and reset-state validity checks and are frozen independently of policy success. For camera and base shifts, the same factor-specific random draw is reused across $L_1$--$L_3$ for each base episode, so severity changes only the perturbation magnitude; illumination and blur are deterministic.
The levels are ordinal within each factor, and equal level indices across different factors do not imply physically equivalent perturbation magnitudes.

\subsection{Counterfactual Paired Evaluation Protocol}
\label{sec:counterfactual_pairing}
To isolate changes associated with a distribution shift from the inherent
variability of RoboCasa episodes, we adopt a paired counterfactual evaluation
protocol. For each base episode $e_i$, we construct a clean rollout
$E_i^{\mathrm{clean}}$ and its perturbed counterpart $E_i^{p,s}$, where $p$
denotes the perturbation factor and $s$ its severity. The paired rollouts share the same task and instruction, scene and object configuration, initial
simulator state, episode horizon, and policy. For stochastic policies, the
same policy-side random seed is initialized for both rollouts in a pair and is not reset at every action step. The target perturbation is therefore the only controlled difference between $E_i^{\mathrm{clean}}$ and $E_i^{p,s}$,
enabling direct comparison under otherwise matched conditions.
We separately validate environment-state identity and policy-side repeatability, with the results summarized in Table~\ref{tab:pairing_validation}.
Implementation details and pairing integrity audits are provided in Appendix~\ref{app:pairing_validation}.

\subsection{Subgoal-Aware Failure Tracking}
\label{sec:stage_tracking}

Episode-level success indicates whether a task is ultimately completed, but
does not reveal how far the policy progresses before failure. We therefore
represent each long-horizon task as an ordered sequence of semantic subgoals,
$\mathcal{G} = (g_1, g_2, \ldots, g_K)$, derived from the task logic and instruction. Each subgoal is associated with a simulator-state predicate that determines whether the corresponding task
condition has been satisfied, such as an object reaching a target receptacle
or a fixture entering a required state. This provides a simulator-grounded
diagnostic of execution progress without requiring external stage annotations
or model-predicted labels.

\begin{table}[t]
\vspace{0.5em}
\centering
\caption{\textbf{Validation of the counterfactual pairing protocol.}
Environment-side pairing is verified through the initial-state fingerprint.
The policy-side comparison shows that uncontrolled inference randomness
breaks repeatability, whereas sharing the episode-level policy seed restores
consistent action sequences.}
\label{tab:pairing_validation}

\small
\renewcommand{\arraystretch}{1.08}
\setlength{\tabcolsep}{3pt}

\begin{tabularx}{\columnwidth}{
@{}
>{\raggedright\arraybackslash}p{0.19\columnwidth}
>{\raggedright\arraybackslash}X
>{\raggedright\arraybackslash}p{0.25\columnwidth}
>{\centering\arraybackslash}p{0.10\columnwidth}
@{}
}
\toprule
\textbf{Component} &
\textbf{Setting} &
\textbf{Validation} &
\textbf{Result} \\
\midrule

Environment &
Shared episode config.\ + seed &
State fingerprint &
\checkmark \\

Policy RNG &
Uncontrolled &
Action sequence &
$\times$ \\

Policy RNG &
Shared episode-level seed &
Action sequence &
\checkmark \\

\bottomrule
\end{tabularx}
\vspace{-2.0em}
\end{table}

\begin{figure}[!b]
    \vspace{-1.5em}
    \centering
    \includegraphics[width=\columnwidth]{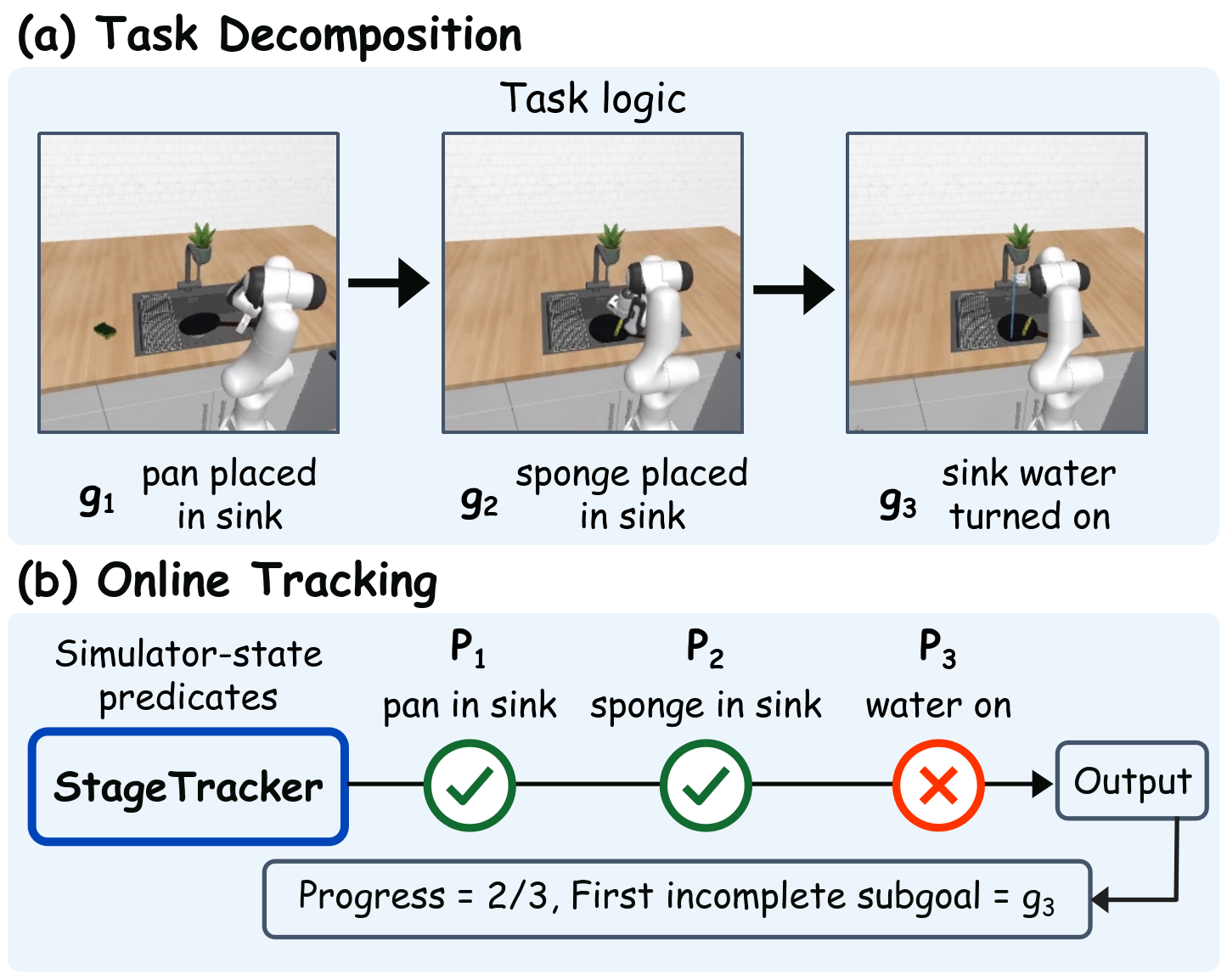}
    \caption{\textbf{Overview of StageTracker.}
    (a) A long-horizon task is represented as an ordered sequence of semantic subgoals, each associated with a simulator-state predicate.
    (b) During rollout, StageTracker evaluates the current predicate, records the completion step when it is first satisfied, and advances to the next subgoal. Completed subgoals are latched, allowing the tracker to report the deepest semantic progress reached and the first incomplete subgoal for failure localization.}
    \label{fig:stage_tracker_overview}
\end{figure}

We implement this sequential tracking procedure with StageTracker,
which evaluates the predicate of the current subgoal at each rollout step. 
Once satisfied, the subgoal is marked as completed, its completion
step is recorded, and the tracker advances to the next one. The tracker also
records the first incomplete subgoal. Completed subgoals are latched rather
than re-evaluated later, so the resulting trajectory represents the deepest
semantic milestone reached during execution rather than the set of conditions
satisfied at the terminal state. StageTracker therefore complements, rather
than replaces, RoboCasa's official task-success criterion by distinguishing
failed executions according to their semantic progress. The procedure is
illustrated in Figure~\ref{fig:stage_tracker_overview}.
Task-specific subgoal specifications and their validation are
provided in Appendix~\ref{app:stage_tracker_validation}.

\vspace{-0.5em}
\subsection{Stage-Aware Robustness Measures}
\label{sec:stage_aware_measures}

RoboCasa's official task-success criterion remains the primary end-to-end
outcome, while StageTracker provides complementary measures of execution
depth. For a task with $K$ ordered subgoals, let $m_i$ denote how many are
sequentially completed in episode $i$. We define the normalized subgoal
progress as
\begin{equation}
\mathrm{Progress}_i = \frac{m_i}{K},
\end{equation}
which measures the deepest normalized semantic milestone reached during
execution. At the task level, for $N$ evaluated episodes, the reach rate of
$g_k$ is
\begin{equation}
R_k = \frac{1}{N}\sum_{i=1}^{N}\mathbb{I}[m_i \ge k],
\end{equation}
representing the fraction of rollouts that reach at least the $k$-th semantic
milestone.

To localize where execution terminates, we further define the first incomplete subgoal
for episodes with $m_i<K$ as
\begin{equation}
f_i = g_{m_i+1}.
\end{equation}
The distribution of $f_i$ provides an operational stage-wise localization of failed executions according to the first unreached semantic milestone. It does not identify the causal instant at which failure is initiated.
Episodes with $m_i=K$ are not assigned
$f_i$, regardless of their final official task-success label. These measures
separate final task completion from the depth and location of execution
failure.
\section{Experiments}

\subsection{Experimental Setup}
\label{sec:experimental_setup}

\textbf{Evaluation Configuration.}
We use the official GR00T N1.5 Composite-Seen checkpoint~\cite{bjorck2025gr00t} as the primary policy, whose training scope matches the target Composite-Seen tasks used in our formal evaluation. All experiments are conducted on the RoboCasa365 target split
\cite{robocasa365} with OSMesa rendering. We evaluate the four clean-qualified tasks introduced in Section~\ref{sec:task_suite}, using 20 fixed base episodes per task. Each episode is evaluated under the clean condition and twelve shifted conditions spanning four perturbation factors and three severity levels, resulting in $20\times13=260$ rollouts per task and 1,040 formal rollouts in total. Each rollout uses the official task horizon, and all clean-perturbed executions follow the counterfactual pairing protocol in Section~\ref{sec:counterfactual_pairing}. A post-run integrity audit confirmed that all formal rollouts passed the base-state, policy-seed, renderer, and StageTracker consistency checks.
Full model/runtime configurations and formal-evaluation integrity statistics
are provided in Appendix~\ref{app:implementation_details}.

\begin{figure}[b]
    \vspace{-1.0em}
    \centering
    \includegraphics[width=\columnwidth]{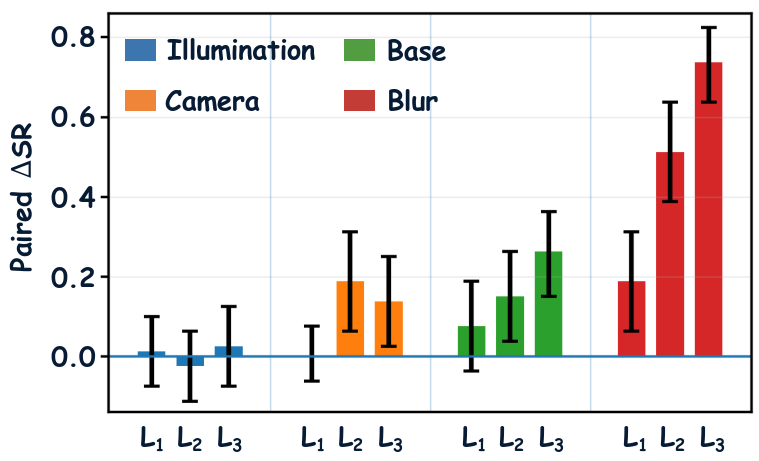}
    \caption{\textbf{End-to-end robustness under controlled distribution shifts.} Macro-averaged paired success-rate degradation $\Delta\mathrm{SR}$ is shown for the four distribution-shift factors, with each factor evaluated at three frozen severity levels ($L_1$--$L_3$). Positive values indicate degradation relative to the matched clean condition. Error bars denote 95\% confidence intervals obtained from paired bootstrap resampling. Severity levels are ordinal within each perturbation factor, and equal level indices across different factors do not imply physically equivalent perturbation magnitudes.}
    \label{fig:rq1_robustness}
\end{figure}

\textbf{Evaluation Metrics and Statistical Analysis.}
We use RoboCasa's official binary task success as the primary end-to-end outcome and report success rate (SR) for each task and evaluation condition.
For perturbation factor $p$ and severity level $s$, paired robustness
degradation is measured as
$\Delta\mathrm{SR}_{p,s}
=\mathrm{SR}_{\mathrm{clean}}-\mathrm{SR}_{p,s}$,
where positive values indicate degradation relative to the matched clean condition. Across tasks, the primary aggregate is the equal-weight macro average of task-level metrics rather than  a pooled estimate over episodes. Confidence intervals for paired degradation are obtained from 10,000 pair-level bootstrap resamples.
Stage-aware analyses use the measures defined in Section~\ref{sec:stage_aware_measures}. For failure localization, we consider only matched pairs in which the clean rollout succeeds and the corresponding perturbed rollout fails, focusing on failures emerging under the evaluated shift.

\begin{figure*}[htbp]
    \vspace{0.5em}
    \centering
    \includegraphics[width=0.9\textwidth]{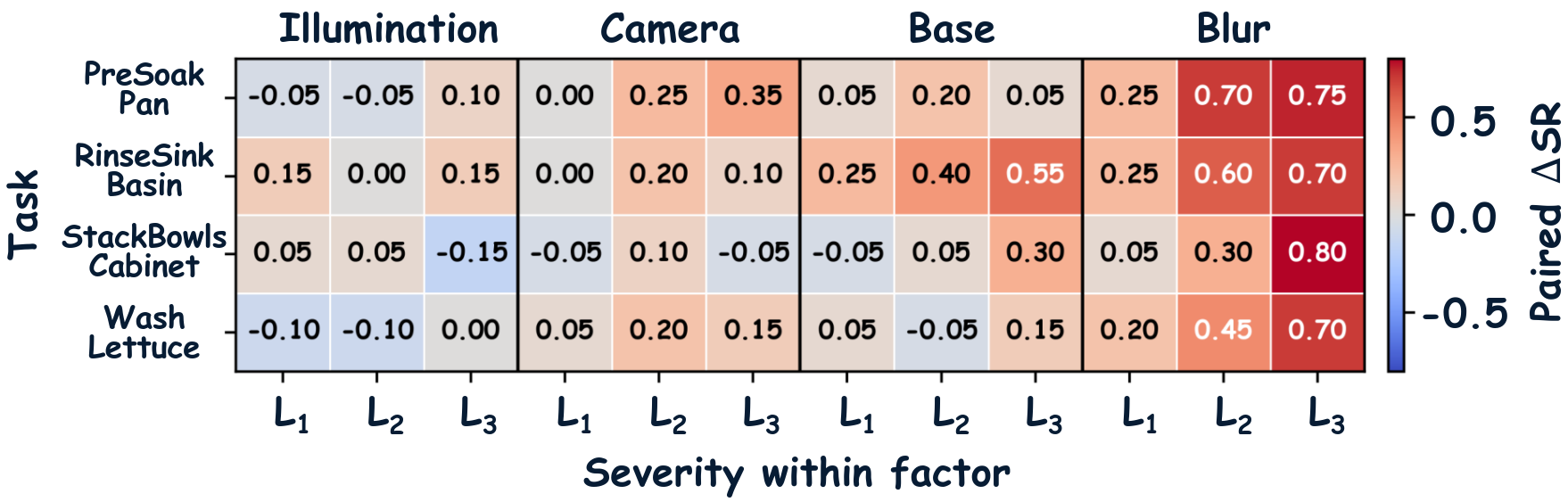}
    \caption{\textbf{Task-level robustness under controlled distribution shifts.} Each cell reports the paired success-rate degradation $\Delta\mathrm{SR}$ for one task and perturbation condition, computed from 20 matched episode pairs. Positive values indicate degradation relative to the matched clean condition, while negative values indicate a higher success rate under the shifted condition for the corresponding finite paired sample. The task-level breakdown reveals heterogeneous severity responses that can be obscured by macro aggregation.}
    \label{fig:rq1_heatmap}
    \vspace{-1.5em}
\end{figure*}

\textbf{Additional Policy Screening.}
We additionally screened two released RoboCasa-compatible VLA checkpoints,
SmolVLA RoboCasa and $\pi_0$ RoboCasa365 Human300, under the same interface
and fixed-seed repeatability checks. Both passed these checks, but achieved
only $1/40$ and $2/40$ clean successes, respectively, with no task satisfying
the predefined competence gate. We therefore did not proceed to cross-model
perturbation evaluation, and the formal perturbation results below are restricted to GR00T N1.5 on the clean-qualified task suite.
Detailed clean-screening outcomes and policy compatibility assessments are provided in
Appendix~\ref{app:cross_model_screening}.


\subsection{End-to-End Robustness under Controlled Distribution Shifts}
\label{sec:results_robustness}

The evaluated policy does not exhibit a universal relationship between perturbation severity and end-to-end robustness. Starting from a clean macro success rate of 0.738, the four distribution-shift factors exhibit distinct severity-response profiles. Moreover, the task-level breakdown shows that similar aggregate trends can arise from heterogeneous responses across individual tasks.
Figure~\ref{fig:rq1_robustness} summarizes the macro paired degradation, while Figure~\ref{fig:rq1_heatmap} shows the corresponding task-level variation.

As shown in Figure~\ref{fig:rq1_robustness}, Gaussian blur exhibits the
clearest and most consistent severity-linked degradation. The macro success
rate decreases from 0.550 at $L_1$ to 0.225 at $L_2$ and reaches 0 at $L_3$,
corresponding to paired $\Delta\mathrm{SR}$ values of 0.188, 0.513, and 0.738,
respectively. The paired 95\% bootstrap confidence intervals exclude zero at
all three severity levels. Robot base initialization also shows a graded
macro-level response, with $\Delta\mathrm{SR}$ increasing from 0.075 at $L_1$
to 0.150 at $L_2$ and 0.263 at $L_3$. The paired confidence intervals exclude
zero at $L_2$ and $L_3$, but not at $L_1$.
The task-level breakdown in Figure~\ref{fig:rq1_heatmap} shows that these
macro trends correspond to different degrees of cross-task consistency.
Gaussian blur produces positive degradation across all four tasks, with
task-level $\Delta\mathrm{SR}$ reaching 0.70--0.80 at $L_3$. In contrast, the
graded macro-level response to robot base initialization is not uniform across
tasks. RinseSinkBasin increases from 0.25 to 0.40 and 0.55 across the three
levels, whereas PreSoakPan changes from 0.05 to 0.20 and then returns to 0.05.
Thus, a graded macro-level response does not necessarily imply a uniformly
monotonic task-level response.

Figure~\ref{fig:rq1_robustness} shows qualitatively different response
patterns for camera orientation and illumination. Camera orientation yields no
macro degradation at $L_1$, followed by $\Delta\mathrm{SR}$ values of 0.188
at $L_2$ and 0.138 at $L_3$, with the paired confidence intervals for the
latter two levels excluding zero. The smaller degradation at $L_3$ than at
$L_2$ shows that increasing calibrated severity does not necessarily translate
into monotonically increasing policy degradation. Illumination produces only
small net macro changes, with $\Delta\mathrm{SR}$ values of 0.012, $-0.025$,
and 0.025, and all three paired confidence intervals include zero.
Figure~\ref{fig:rq1_heatmap} further reveals heterogeneous task-level
responses underlying these aggregate patterns. For camera orientation,
PreSoakPan degrades progressively from 0.00 to 0.25 and 0.35, while the other
tasks show smaller degradation at $L_3$ than at $L_2$ or changes in the
opposite direction. Under illumination, task-level changes also occur in both
directions, indicating that a near-zero macro effect can coexist with
heterogeneous task responses rather than uniformly unchanged behavior.
End-to-end robustness therefore varies across perturbations and tasks, while
observed severity responses are not uniformly monotonic.
Complete condition-level numerical results are provided in Appendix~\ref{app:rq1_results}.


\subsection{Counterfactual Sensitivity Beyond Aggregate Success}
\label{sec:counterfactual_sensitivity}

Aggregate success-rate changes can mask substantial episode-level sensitivity to a distribution shift. Under strict clean-perturbed pairing, each matched episode can be classified according to its clean and perturbed outcomes as
success-to-success ($\mathrm{S}\!\rightarrow\!\mathrm{S}$),
success-to-failure ($\mathrm{S}\!\rightarrow\!\mathrm{F}$),
failure-to-success ($\mathrm{F}\!\rightarrow\!\mathrm{S}$), or
failure-to-failure ($\mathrm{F}\!\rightarrow\!\mathrm{F}$), where the first and second symbols denote the clean and perturbed outcomes, respectively.
The aggregate success-rate drop depends only on the net difference between the two outcome-changing transitions,
\begin{equation}
\Delta \mathrm{SR}
=
\frac{N_{\mathrm{S}\rightarrow\mathrm{F}}
-
N_{\mathrm{F}\rightarrow\mathrm{S}}}{N},
\label{eq:transition_delta_sr}
\end{equation}
so opposing transition directions can partially or completely cancel at the aggregate level.

\begin{table}[!b]
\vspace{-1.0em}
\centering
\caption{\textbf{Counterfactual outcome transitions across all perturbation
conditions.} Each condition contains 80 strictly matched clean--perturbed
episode pairs. Changed denotes the number of pairs whose binary task outcome
differs between the clean and perturbed executions.}
\label{tab:counterfactual_transitions}
\small
\renewcommand{\arraystretch}{1.05}

\begin{tabular*}{\columnwidth}{
@{\extracolsep{\fill}}
lccccc
@{}
}
\toprule
\textbf{Factor} &
\textbf{Level} &
$\boldsymbol{\Delta\mathrm{SR}}$ &
$\mathbf{S}\!\rightarrow\!\mathbf{F}$ &
$\mathbf{F}\!\rightarrow\!\mathbf{S}$ &
\textbf{Changed} \\
\midrule

Illumination
& $L_1$ & 0.012  & 7  & 6 & 13 \\
& $L_2$ & -0.025 & 5  & 7 & 12 \\
& $L_3$ & 0.025  & 10 & 8 & 18 \\

\midrule
Camera
& $L_1$ & 0.000 & 4  & 4 & 8 \\
& $L_2$ & 0.188 & 21 & 6 & 27 \\
& $L_3$ & 0.138 & 18 & 7 & 25 \\

\midrule
Base
& $L_1$ & 0.075 & 14 & 8 & 22 \\
& $L_2$ & 0.150 & 19 & 7 & 26 \\
& $L_3$ & 0.263 & 24 & 3 & 27 \\

\midrule
Gaussian Blur
& $L_1$ & 0.188 & 22 & 7 & 29 \\
& $L_2$ & 0.513 & 44 & 3 & 47 \\
& $L_3$ & 0.738 & 59 & 0 & 59 \\

\bottomrule
\end{tabular*}
\end{table}

Table~\ref{tab:counterfactual_transitions} summarizes the paired outcome
transitions across all twelve perturbation conditions. The three cases below highlight distinct relationships between aggregate degradation and pair-level outcome changes.
Under camera orientation at $L_1$, $\Delta\mathrm{SR}=0.000$ despite 8 of the 80 matched episodes changing outcome, as four $\mathrm{S}\!\rightarrow\!\mathrm{F}$ and four $\mathrm{F}\!\rightarrow\!\mathrm{S}$ transitions cancel exactly.
Illumination at $L_2$ shows a similar mismatch, with $\Delta\mathrm{SR}=-0.025$ despite 12 matched episodes changing outcome. In contrast, Gaussian blur at $L_3$ produces 59 $\mathrm{S}\!\rightarrow\!\mathrm{F}$ transitions and no $\mathrm{F}\!\rightarrow\!\mathrm{S}$ transitions, yielding $\Delta\mathrm{SR}=0.738$. In this case, the pair-level outcome changes are strongly aligned with the aggregate degradation. 

These results distinguish net robustness degradation from counterfactual episode sensitivity. A small $\Delta\mathrm{SR}$ does not imply that the underlying matched executions are insensitive to the shift, because $\mathrm{S}\!\rightarrow\!\mathrm{F}$ and
$\mathrm{F}\!\rightarrow\!\mathrm{S}$ transitions can offset each other in the aggregate.
Moreover, $\mathrm{F}\!\rightarrow\!\mathrm{S}$ transitions should not be interpreted as evidence that a perturbation improves robustness. They indicate only that the matched episode outcome changes under the shifted condition.
Complete four-way outcome statistics and representative
qualitative examples are provided in
Appendices~\ref{app:counterfactual_transitions}
and~\ref{app:qualitative_counterfactual} respectively.

\begin{figure}[b]
    \vspace{-1.0em}
    \centering
    \includegraphics[width=\columnwidth]{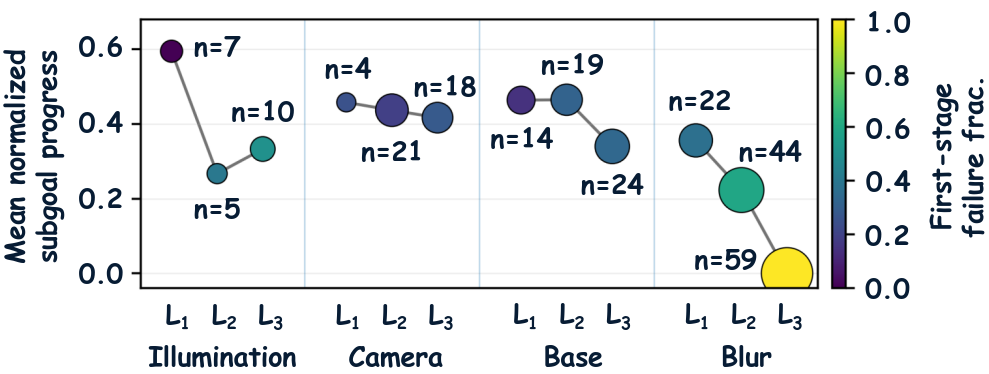}
    \caption{\textbf{Stage-aware execution depth of perturbation-induced failures.}
    Each marker summarizes strictly matched clean-success-to-perturbed-failure ($\mathrm{S}\!\rightarrow\!\mathrm{F}$) transitions under one perturbation factor and severity level. The vertical position indicates the mean
    normalized subgoal progress of the corresponding
    $\mathrm{S}\!\rightarrow\!\mathrm{F}$ cohort, marker size represents the number of transition events, and marker color denotes the fraction of failures occurring before the first tracked semantic milestone. The
    annotated $n$ gives the number of transitions in each condition. Gray lines connect $L_1$--$L_3$ within the same perturbation factor only to visualize the within-factor severity trend and do not encode an additional
    quantity. Severity levels are ordinal within each perturbation factor and are not assumed to be physically equivalent across factors.}
    \label{fig:rq2_failure_frontier}
\end{figure}

\begin{figure*}[htbp]
    \vspace{0.5em}
    \centering
    \includegraphics[width=\textwidth]{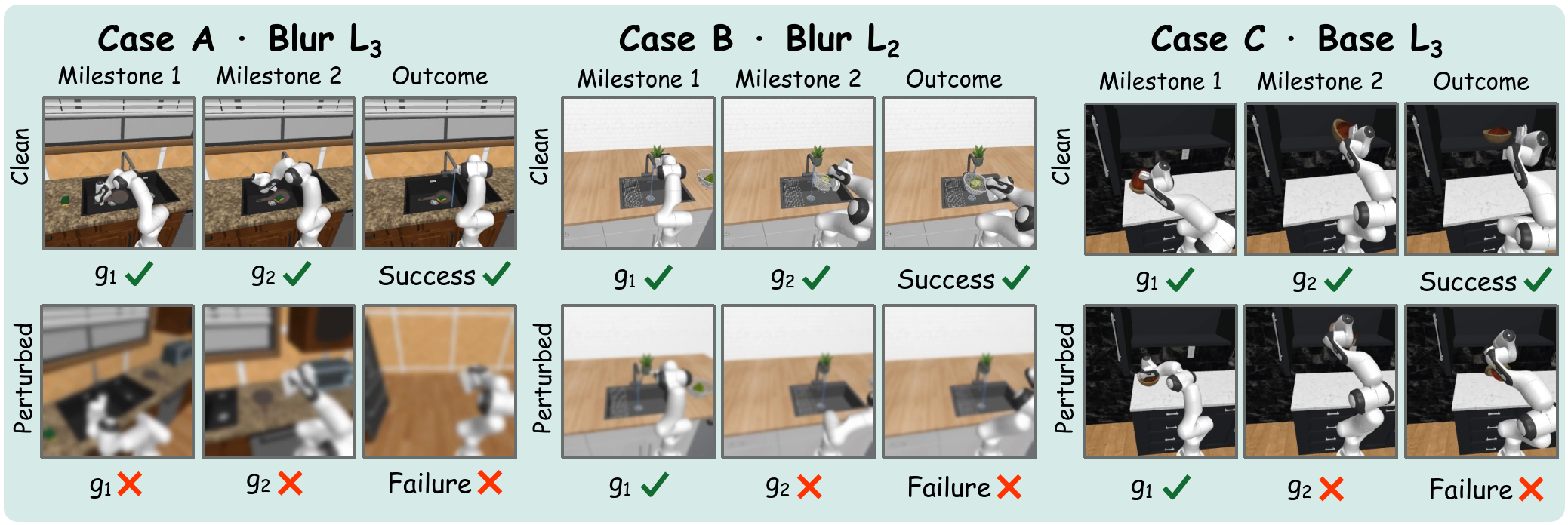}
    \caption{\textbf{Representative matched replays at different semantic failure depths.}
    Each case compares a strictly matched clean--perturbed rollout pair that shares the same task instance, initial simulator state, episode metadata,
    and episode-level policy seed, while differing only in the target perturbation. The two rows show the clean and perturbed executions, respectively, and the columns present selected frames spanning the execution process. 
    Checkmarks and crosses indicate whether the corresponding tracked semantic subgoals are completed, while the final annotation reports the official task outcome. These examples provide a qualitative view of how perturbations can change execution depth and
    failure location, complementing the aggregate stage-aware statistics in
    Figure~\ref{fig:rq2_failure_frontier}.}
    \label{fig:qualitative_replay}
    \vspace{-1.5em}
\end{figure*}

\subsection{Semantic Depth of Perturbation-Induced Failures}
\label{sec:results_failure_localization}

Perturbation-induced failure episodes reach different semantic depths of task execution, depending on perturbation factor and severity. We restrict this analysis to strictly matched
$\mathrm{S}\!\rightarrow\!\mathrm{F}$ pairs. Across the twelve factor-severity conditions, we observe 247 condition-specific paired transitions in total. We characterize each cohort using normalized
subgoal progress and the first incomplete semantic milestone, as summarized in Figure~\ref{fig:rq2_failure_frontier}.
Complete condition-level statistics and task-wise milestone distributions are provided in Appendix~\ref{app:rq2_results}.

Similar end-to-end degradation can nevertheless correspond to different semantic execution depths. Camera orientation at $L_2$ and Gaussian blur at $L_1$ produce the same macro degradation of
$\Delta\mathrm{SR}=0.188$, yet exhibit different semantic
execution-depth profiles. The camera-orientation cohort reaches a mean normalized progress of 0.437, with 19.0\% of failures occurring before the first tracked milestone. Under Gaussian blur, the corresponding mean progress is lower at 0.356, while the first-stage failure fraction increases to 36.4\%. Thus, comparable changes in final task success can correspond to different semantic depths that end-to-end success alone cannot distinguish.



Gaussian blur also exhibits the clearest severity-dependent shift in
stage-aware execution depth. The number of $\mathrm{S}\!\rightarrow\!\mathrm{F}$ events
increases from 22 at $L_1$ to 44 at $L_2$ and 59 at $L_3$, while the mean normalized subgoal progress decreases from 0.356 to 0.223 and finally to 0.000. Over the same severity range, the fraction of failures occurring before the first tracked milestone rises from 36.4\% to 59.1\% and then to 100\%.
At $L_3$, all 59 perturbation-induced failures terminate before completing the first tracked semantic milestone. Stronger blur therefore affects not only how frequently clean-successful episodes fail under perturbation, but also how little semantic progress the resulting failure episodes achieve.


The remaining perturbation factors exhibit weaker or less uniform localization patterns. For robot base initialization, mean normalized progress is 0.464, 0.465, and 0.340 from $L_1$ to $L_3$, while the first-stage failure fraction increases from 14.3\% to 31.6\% and 33.3\%. This suggests a tendency toward earlier failure under stronger base shifts, although the progression is not
uniform across severity levels. Camera orientation produces failures across multiple execution depths, with mean progress decreasing modestly from 0.458 to 0.437 and 0.417, but without a consistent monotonic change in the first-incomplete distribution. Illumination yields comparatively few $\mathrm{S}\!\rightarrow\!\mathrm{F}$ events and does not show a stable failure-localization pattern over the evaluated severity range. The localization response therefore remains perturbation dependent rather than following a universal severity trend. Representative matched replays in Figure~\ref{fig:qualitative_replay} illustrate these differences,  with perturbed executions terminating at different semantic depths under otherwise matched rollout conditions.

\vspace{-0.2em}

\subsection{Scope and Limitations}
\label{sec:scope_limitations}

The formal perturbation results are obtained with GR00T N1.5 on four
long-horizon RoboCasa tasks selected by the predefined clean-performance
gate. Because no additional screened policy passed this gate, we do not claim
cross-model generalization. The four-task suite also does not support
controlled claims across broader task structures or mobility settings. Our
stage-aware diagnostics rely on audited simulator-state predicates rather than
official per-frame stage annotations and should therefore be interpreted as
execution diagnostics rather than ground-truth stage labels. Finally, all
evaluations are conducted in simulation, and whether the observed robustness
patterns transfer to real-robot deployment remains to be validated. Extending
the protocol to stronger policies, broader task coverage, and real-robot
settings is an important direction for future work.



\section{Conclusion}
\label{sec:conclusion}


We presented \textbf{RoboCasa-Shift}, a diagnostic robustness benchmark that combines matched clean-perturbed evaluation with simulator-grounded stage-aware diagnostics for long-horizon household manipulation. Our experiments show that robustness varies across perturbations and tasks, while aggregate success can mask substantial episode-level sensitivity. Similar end-to-end degradation can also arise from failures at different semantic depths, with stronger shifts in some conditions associated with earlier incomplete milestones. These findings motivate moving beyond endpoint-only robustness reporting toward evaluation protocols that diagnose how distribution shifts
affect task execution.






\bibliographystyle{ieeetr}
\bibliography{references} 

\begin{table*}[!b]
    \centering
    \caption{\textbf{Complete clean-competence screening results.}
    The screening rule and eligibility threshold are defined in
    Section~\ref{sec:task_suite}. Tasks not extended beyond the initial
    10 episodes are clean-floor cases.}
    \label{tab:clean_screening}
    \small
    \setlength{\tabcolsep}{5pt}
    \begin{tabular}{lcccccc}
        \toprule
        Task &
        Initial &
        Final $N$ &
        Success &
        Clean SR &
        Wilson 95\% CI &
        Status \\
        \midrule
        DeliverStraw
            & 2/10  & 10 & 2  & 0.20 & [0.057, 0.510] & Floor \\
        GetToastedBread
            & 0/10  & 10 & 0  & 0.00 & [0.000, 0.278] & Floor \\
        KettleBoiling
            & 5/10  & 20 & 9  & 0.45 & [0.258, 0.658] & Below gate \\
        LoadDishwasher
            & 5/10  & 20 & 9  & 0.45 & [0.258, 0.658] & Below gate \\
        PackIdenticalLunches
            & 1/10  & 10 & 1  & 0.10 & [0.018, 0.404] & Floor \\
        PreSoakPan
            & 7/10  & 20 & 15 & 0.75 & [0.531, 0.888] & Eligible \\
        PrepareCoffee
            & 2/10  & 10 & 2  & 0.20 & [0.057, 0.510] & Floor \\
        RinseSinkBasin
            & 10/10 & 20 & 14 & 0.70 & [0.481, 0.855] & Eligible \\
        ScrubCuttingBoard
            & 4/10  & 20 & 10 & 0.50 & [0.299, 0.701] & Below gate \\
        SearingMeat
            & 4/10  & 20 & 7  & 0.35 & [0.181, 0.567] & Below gate \\
        SetUpCuttingStation
            & 0/10  & 10 & 0  & 0.00 & [0.000, 0.278] & Floor \\
        StackBowlsCabinet
            & 7/10  & 20 & 16 & 0.80 & [0.584, 0.919] & Eligible \\
        SteamInMicrowave
            & 5/10  & 20 & 11 & 0.55 & [0.342, 0.742] & Below gate \\
        StirVegetables
            & 2/10  & 10 & 2  & 0.20 & [0.057, 0.510] & Floor \\
        StoreLeftoversInBowl
            & 6/10  & 20 & 10 & 0.50 & [0.299, 0.701] & Below gate \\
        WashLettuce
            & 8/10  & 20 & 14 & 0.70 & [0.481, 0.855] & Eligible \\
        \bottomrule
    \end{tabular}
\end{table*}

\section{Appendix}
\subsection{Task Selection and Clean-Competence Screening}
\label{app:task_selection}

The complete candidate-level clean-screening results are reported in
Table~\ref{tab:clean_screening}. Although the 16-task candidate pool was
frozen in advance, screening was executed in a preregistered $12{+}4$
schedule. The first 12 candidates yielded three clean-qualified tasks; because
this was below the predefined minimum of four formal tasks, the remaining four
pre-frozen candidates were activated. Among them, only \textit{WashLettuce}
passed the same clean-performance gate, producing the final four-task suite.

\subsection{Perturbation Implementation and Severity Calibration}
\label{app:perturbation_details}

\subsubsection{Perturbation Implementation}
The perturbations are implemented directly within the simulator or observation
pipeline while keeping the underlying task configuration unchanged. Each
perturbation modifies only the targeted factor, without altering task
semantics, object placement, or language instructions.

For illumination shifts, we modify the simulator lighting parameters by scaling
the diffuse and ambient components. For camera-orientation shifts, we perturb
the external camera pose while keeping wrist-mounted views unchanged. The
camera transformation is applied before rendering, producing observation-level
viewpoint changes without modifying the underlying scene geometry.
For robot base initialization shifts, Gaussian offsets are sampled for the
mobile base planar position and yaw orientation before rollout reset. The perturbation magnitude increases
with severity while maintaining valid initial states. For Gaussian blur, the
corruption is applied to RGB observations from all relevant camera streams using
Gaussian filtering with increasing kernel scales. These implementations allow
each factor to be evaluated independently while preserving paired
clean-perturbed comparisons.

\subsubsection{Severity Calibration}
\begin{figure*}[!t]
    \centering
    \includegraphics[width=0.96\textwidth]{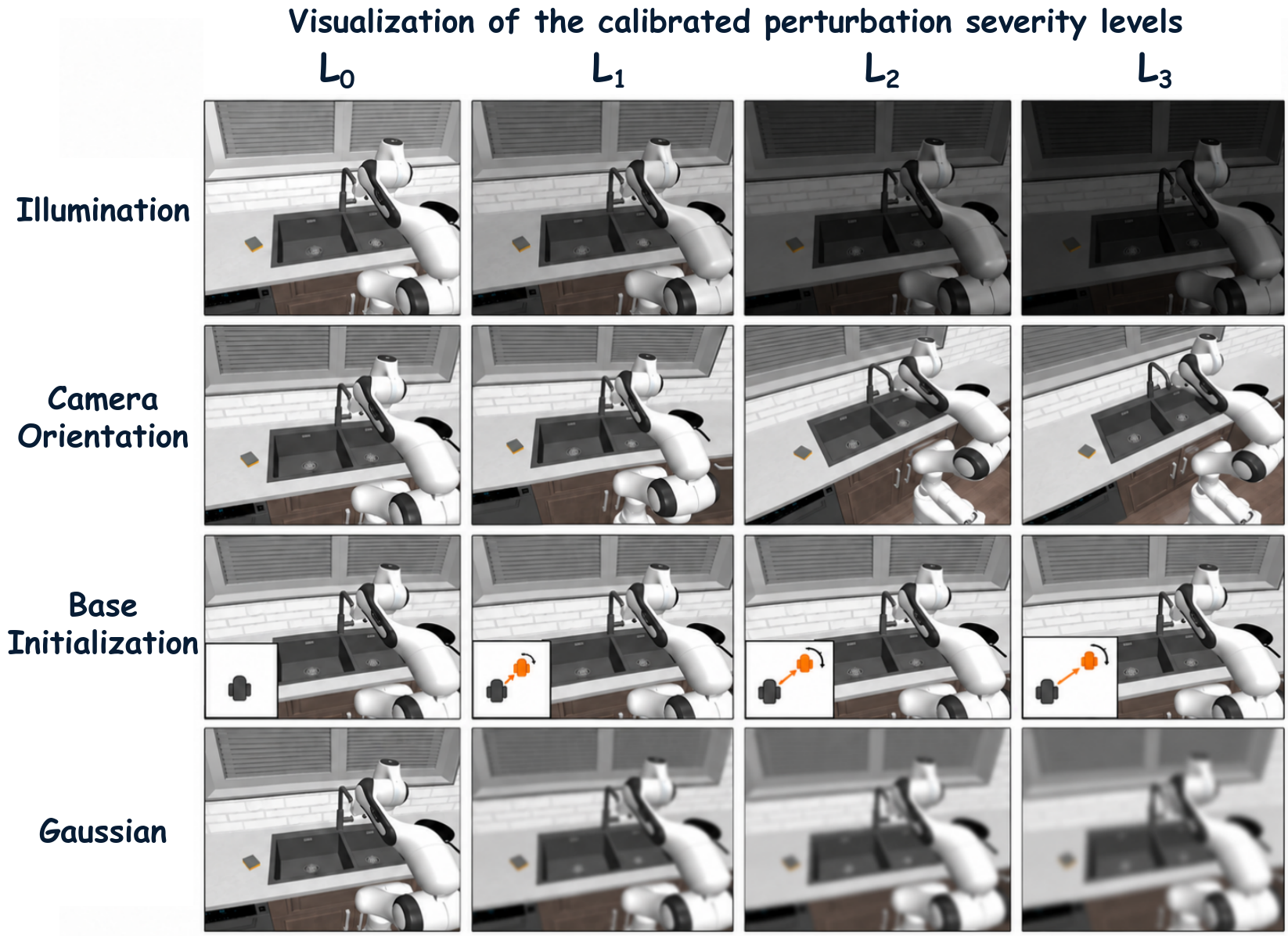}
   \caption{\textbf{Representative severity calibration examples.}
    Rows correspond to the four perturbation factors, and columns show the clean condition ($L_0$) and three calibrated severity levels ($L_1$--$L_3$).
    Illumination progressively darkens the scene, camera orientation
    changes the third-person viewpoint, robot base initialization
    perturbs the initial base position and yaw orientation, and
    Gaussian blur progressively reduces visual sharpness.
    For base initialization, the inset provides a schematic top-down
    view, where the black and orange markers represent the nominal
    and perturbed base poses, respectively, while the straight and
    curved arrows indicate planar displacement and yaw rotation.
    Severity levels are ordinal within each perturbation factor
    and are not assumed to be physically equivalent across factors.}
    \label{fig:severity_visualization}
    \vspace{-0.5em}
\end{figure*}

Before formal evaluation, we conducted a model-free severity validity audit
on the four formal tasks, using three fixed base episodes per task under
the clean condition and twelve perturbation conditions. This yielded
156 reset/render records, which were checked for valid initialization,
observation integrity, pre-perturbation state consistency, and the
intended factor-specific changes.
All 156 records passed the automatic validity checks, with no reported
errors or pre-perturbation base-state fingerprint mismatches. Illumination
and Gaussian blur showed the expected severity-dependent changes without
monotonicity warnings, and manual contact-sheet inspection also passed.
No policy was evaluated during this audit, ensuring that severity validation
was independent of policy success. These checks establish the validity
of the audited reset and rendering conditions, rather than guaranteeing
task completion under perturbation.
Representative examples of the calibrated severity levels are shown in
Figure~\ref{fig:severity_visualization}. 

\subsection{Counterfactual Pairing Validation}
\label{app:pairing_validation}

\subsubsection{Pair Construction}

Each perturbed rollout is generated from the corresponding clean episode
configuration while preserving the matched task conditions and policy-side
stochasticity. The target perturbation is applied before execution, while the
task objective and evaluation protocol remain unchanged. This construction
controls episode-level variation and enables analysis of
perturbation-induced outcome changes.

\subsubsection{Pairing Integrity Audit}

We validate pairing consistency from both environment and policy perspectives.
For environment-side consistency, we compare simulator state fingerprints
before perturbation injection between clean and perturbed executions. Matching
fingerprints indicate that the two executions share the same pre-shift
simulator configuration.
For policy-side consistency, we restore the same policy checkpoint and random
number generator state before paired execution. This controls stochastic
variation introduced by policy sampling without assuming deterministic action
generation.
The final audit verifies that all paired executions satisfy the required
consistency conditions before formal evaluation. The validation criteria are
summarized in Table~\ref{tab:pairing_audit}.

\begin{table}[H]
    \centering
    \caption{\textbf{Pairing integrity validation.}
    Environment-side checks verify consistency of the simulator configuration
    before perturbation injection, while policy-side checks control stochastic
    variation during paired execution.}
    \label{tab:pairing_audit}
    \small
    \setlength{\tabcolsep}{5pt}
    \begin{tabular}{lll}
        \toprule
        Validation &
        Procedure &
        Result \\
        \midrule
        Environment state &
        Pre-shift simulator fingerprint match &
        Pass \\
        
        Policy stochasticity &
        Shared checkpoint and RNG state &
        Pass \\
        
        Rollout metadata &
        Matched episode configuration &
        Pass \\
        \bottomrule
    \end{tabular}
\end{table}

\subsection{StageTracker Specifications and Validation}
\label{app:stage_tracker_validation}

\subsubsection{Task-Specific Subgoal Specifications}
Table~\ref{tab:task_subgoal_specifications} lists the ordered semantic
milestones and their simulator-state completion criteria for the four
formal evaluation tasks.

\begin{table}[H]
    \centering
    \caption{\textbf{Task-specific semantic subgoal specifications.}
    Each $g_k$ denotes a task-specific diagnostic milestone.
    The criteria were frozen before formal perturbation evaluation.
    For WashLettuce, thresholds refer to the simulator
    \texttt{washed\_time} counter.}
    \label{tab:task_subgoal_specifications}
    \footnotesize
    \setlength{\tabcolsep}{3pt}
    \renewcommand{\arraystretch}{1.12}
    \begin{tabular}{@{}lcp{0.49\columnwidth}@{}}
        \toprule
        Task & Milestone & Completion criterion \\
        \midrule
        PreSoakPan
            & $g_1$ & Pan in sink; \\
            & $g_2$ & Sponge in sink. \\
            & $g_3$ & Sink water is on. \\
        \midrule
        RinseSinkBasin
            & $g_1$ & Sink water is on. \\
            & $g_2$ & At least two basin zones rinsed. \\
            & $g_3$ & All three basin zones rinsed. \\
        \midrule
        StackBowlsCabinet
            & $g_1$ & One bowl inside the other. \\
            & $g_2$ & Both bowls in cabinet; gripper away from both. \\
        \midrule
        WashLettuce
            & $g_1$ & Sink water is on. \\
            & $g_2$ & $\texttt{washed\_time} \geq 1$. \\
            & $g_3$ & $\texttt{washed\_time} \geq 25$. \\
        \bottomrule
    \end{tabular}
\end{table}

\subsubsection{StageTracker Validation}
Before formal evaluation, the frozen task specifications were audited
by replaying five successful target demonstrations per task through
StageTracker using recorded simulator states. All 20 demonstrations
reached their specified milestones in the expected order, with no
predicate-evaluation errors. This audit establishes consistency of
the task-specific predicates on the evaluated demonstrations, rather
than agreement with ground-truth per-frame stage annotations.

\subsection{Implementation Details}
\label{app:implementation_details}

\paragraph{Model and Runtime Configuration}
The complete model, inference, software, and hardware configurations
used in the formal evaluation are summarized in
Table~\ref{tab:implementation_details}. To accommodate the Blackwell GPU,
we use the scaled dot-product attention (SDPA) backend instead of the
original FlashAttention path, without modifying the policy checkpoint.

\begin{table}[!b]
    \centering
    \caption{\textbf{Integrity validation of the formal evaluation.}
    Each task contains 20 fixed base episodes evaluated under 13
    conditions, resulting in 260 expected rollouts per task.}
    \label{tab:formal_protocol_validation}
    \small
    \setlength{\tabcolsep}{3.5pt}
    \resizebox{\columnwidth}{!}{
    \begin{tabular}{lrrrrrc}
        \toprule
        Task &
        Expected &
        Valid &
        Pairs &
        Hash Mismatch &
        Seed Mismatch &
        Status \\
        \midrule
        PreSoakPan
            & 260 & 260 & 20 & 0 & 0 & PASS \\
        RinseSinkBasin
            & 260 & 260 & 20 & 0 & 0 & PASS \\
        StackBowlsCabinet
            & 260 & 260 & 20 & 0 & 0 & PASS \\
        WashLettuce
            & 260 & 260 & 20 & 0 & 0 & PASS \\
        \bottomrule
    \end{tabular}}
\end{table}

\paragraph{Protocol Validation}
We perform a post-run integrity audit of the complete formal evaluation.
As shown in Table~\ref{tab:formal_protocol_validation}, all four tasks
contain the expected 260 valid rollouts across 20 fixed base episode IDs,
yielding 1,040 valid rollouts in total. All task-level validators pass,
with no pre-perturbation base-state hash mismatches or policy-seed mismatches.
The audit also reports no renderer errors, invalid perturbation runtime
records, runtime failures, StageTracker errors, or unexpected truncations.

\begin{table}[t]
    \centering
    \caption{\textbf{Complete macro-level end-to-end robustness results.}
    Success rate (SR) and paired success-rate degradation
    $\Delta\mathrm{SR}$ are averaged equally across the four formal tasks.
    Positive $\Delta\mathrm{SR}$ indicates degradation relative to the
    matched clean condition. Confidence intervals are obtained from
    paired bootstrap resampling.}
    \label{tab:rq1_full_results}
    \small
    \setlength{\tabcolsep}{5pt}
    \resizebox{\columnwidth}{!}{
    \begin{tabular}{llccc}
        \toprule
        Factor &
        Level &
        SR &
        $\Delta\mathrm{SR}$ &
        95\% CI of $\Delta\mathrm{SR}$ \\
        \midrule

        Clean
        & $L_0$
        & 0.738
        & 0.000
        & [0.000, 0.000] \\

        \midrule

        Illumination
        & $L_1$
        & 0.725
        & 0.012
        & [-0.075, 0.100] \\

        Illumination
        & $L_2$
        & 0.762
        & -0.025
        & [-0.112, 0.062] \\

        Illumination
        & $L_3$
        & 0.713
        & 0.025
        & [-0.075, 0.125] \\

        \midrule

        Camera Orientation
        & $L_1$
        & 0.738
        & 0.000
        & [-0.063, 0.075] \\

        Camera Orientation
        & $L_2$
        & 0.550
        & 0.188
        & [0.063, 0.313] \\

        Camera Orientation
        & $L_3$
        & 0.600
        & 0.138
        & [0.025, 0.250] \\

        \midrule

        Base Initialization
        & $L_1$
        & 0.663
        & 0.075
        & [-0.038, 0.188] \\

        Base Initialization
        & $L_2$
        & 0.588
        & 0.150
        & [0.038, 0.263] \\

        Base Initialization
        & $L_3$
        & 0.475
        & 0.263
        & [0.150, 0.363] \\

        \midrule

        Gaussian Blur
        & $L_1$
        & 0.550
        & 0.188
        & [0.063, 0.313] \\

        Gaussian Blur
        & $L_2$
        & 0.225
        & 0.513
        & [0.388, 0.638] \\

        Gaussian Blur
        & $L_3$
        & 0.000
        & 0.738
        & [0.638, 0.825] \\

        \bottomrule
    \end{tabular}}
\end{table}

\subsection{Additional End-to-End Robustness Results}
\label{app:rq1_results}

Table~\ref{tab:rq1_full_results} provides the complete macro-level
numerical results corresponding to the end-to-end robustness analysis
in Section~\ref{sec:results_robustness}, including success rates,
paired success-rate degradation, and the corresponding 95\%
paired-bootstrap confidence intervals for all perturbation conditions.

\begin{table}[b]
    \centering
    \caption{\textbf{Complete counterfactual outcome transitions across
    all perturbation conditions.} Each row contains 80 strictly matched
    clean--perturbed episode pairs, categorized into four mutually
    exclusive outcomes: success-to-success
    ($\mathrm{S}\!\rightarrow\!\mathrm{S}$),
    success-to-failure ($\mathrm{S}\!\rightarrow\!\mathrm{F}$),
    failure-to-success ($\mathrm{F}\!\rightarrow\!\mathrm{S}$), and
    failure-to-failure ($\mathrm{F}\!\rightarrow\!\mathrm{F}$).}
    \label{tab:counterfactual_transitions_full}
    \small
    \setlength{\tabcolsep}{5pt}
    \resizebox{\columnwidth}{!}{
    \begin{tabular}{llrrrr}
        \toprule
        Factor &
        Level &
        $\mathrm{S}\!\rightarrow\!\mathrm{S}$ &
        $\mathrm{S}\!\rightarrow\!\mathrm{F}$ &
        $\mathrm{F}\!\rightarrow\!\mathrm{S}$ &
        $\mathrm{F}\!\rightarrow\!\mathrm{F}$ \\
        \midrule

        Illumination
        & $L_1$ & 52 & 7  & 6 & 15 \\
        & $L_2$ & 54 & 5  & 7 & 14 \\
        & $L_3$ & 49 & 10 & 8 & 13 \\

        \midrule

        Camera Orientation
        & $L_1$ & 55 & 4  & 4 & 17 \\
        & $L_2$ & 38 & 21 & 6 & 15 \\
        & $L_3$ & 41 & 18 & 7 & 14 \\

        \midrule

        Base Initialization
        & $L_1$ & 45 & 14 & 8 & 13 \\
        & $L_2$ & 40 & 19 & 7 & 14 \\
        & $L_3$ & 35 & 24 & 3 & 18 \\

        \midrule

        Gaussian Blur
        & $L_1$ & 37 & 22 & 7 & 14 \\
        & $L_2$ & 15 & 44 & 3 & 18 \\
        & $L_3$ & 0  & 59 & 0 & 21 \\

        \bottomrule
    \end{tabular}}
\end{table}

\begin{table*}[t]
    \centering
    \caption{\textbf{Model, inference, and runtime configuration used in the
    formal evaluation.} The left half summarizes the GR00T policy and
    evaluation settings, while the right half reports the software and
    hardware environment. Runtime package versions are reported separately
    from the underlying container configuration.}
    \label{tab:implementation_details}
    \footnotesize
    \setlength{\tabcolsep}{4pt}
    \renewcommand{\arraystretch}{1.08}
    \begin{tabular}{
        @{}
        p{0.20\textwidth}
        p{0.28\textwidth}
        p{0.20\textwidth}
        p{0.28\textwidth}
        @{}
    }
        \toprule
        \multicolumn{2}{c}{\textbf{Model and Inference}} &
        \multicolumn{2}{c}{\textbf{Runtime Environment}} \\
        \cmidrule(r){1-2}
        \cmidrule(l){3-4}
        Configuration & Value &
        Configuration & Value \\
        \midrule

        Policy &
        GR00T N1.5 &
        Base container &
        \texttt{nvidia24.04-ubuntu22.04} \\

        Checkpoint &
        \texttt{composite\_seen/checkpoint-60000} &
        Base Python / CUDA &
        Python 3.10 / CUDA 12.4 \\

        Evaluation split &
        Target &
        Base cuDNN / PyTorch &
        cuDNN 9.1 / PyTorch 2.3.0 \\

        Training / task scope &
        Composite-Seen &
        GPU &
        NVIDIA GeForce RTX 5090 \\

        Data configuration &
        \texttt{panda\_omron} &
        GPU memory &
        32607 MiB \\

        Embodiment tag &
        \texttt{new\_embodiment} &
        NVIDIA driver &
        580.95.05 \\

        Denoising steps &
        4 &
        Runtime Python &
        3.10.12 \\

        Executed action steps / prediction &
        16 &
        Runtime PyTorch &
        2.7.1+cu128 \\

        RGB observations &
        Two third-person + wrist &
        CUDA runtime &
        12.8 \\

        Attention backend &
        SDPA &
        MuJoCo / RoboCasa &
        3.3.1 / 1.0.1 \\

        Renderer &
        OSMesa &
        robosuite / robomimic &
        1.5.2 / 0.5.0 \\

        Master seed &
        20260820 &
        GR00T / Transformers &
        1.1.0 / 4.51.3 \\

        Policy-seed base &
        17082020 &
        NumPy / Diffusers &
        1.26.4 / 0.30.2 \\

        Policy randomness &
        Fixed once per episode &
        & \\

        \bottomrule
    \end{tabular}
\end{table*}

\subsection{Complete Counterfactual Transition Statistics}
\label{app:counterfactual_transitions}

Table~\ref{tab:counterfactual_transitions_full} reports the complete
four-way paired outcome counts across all twelve perturbation conditions.
These statistics complement the outcome-changing transitions presented
in Section~\ref{sec:counterfactual_sensitivity} by additionally accounting
for pairs whose success or failure outcomes remain unchanged.

\subsection{Additional Stage-Aware Failure Analysis}
\label{app:rq2_results}

\begin{figure*}[t]
    \centering
    \includegraphics[width=\textwidth]
    {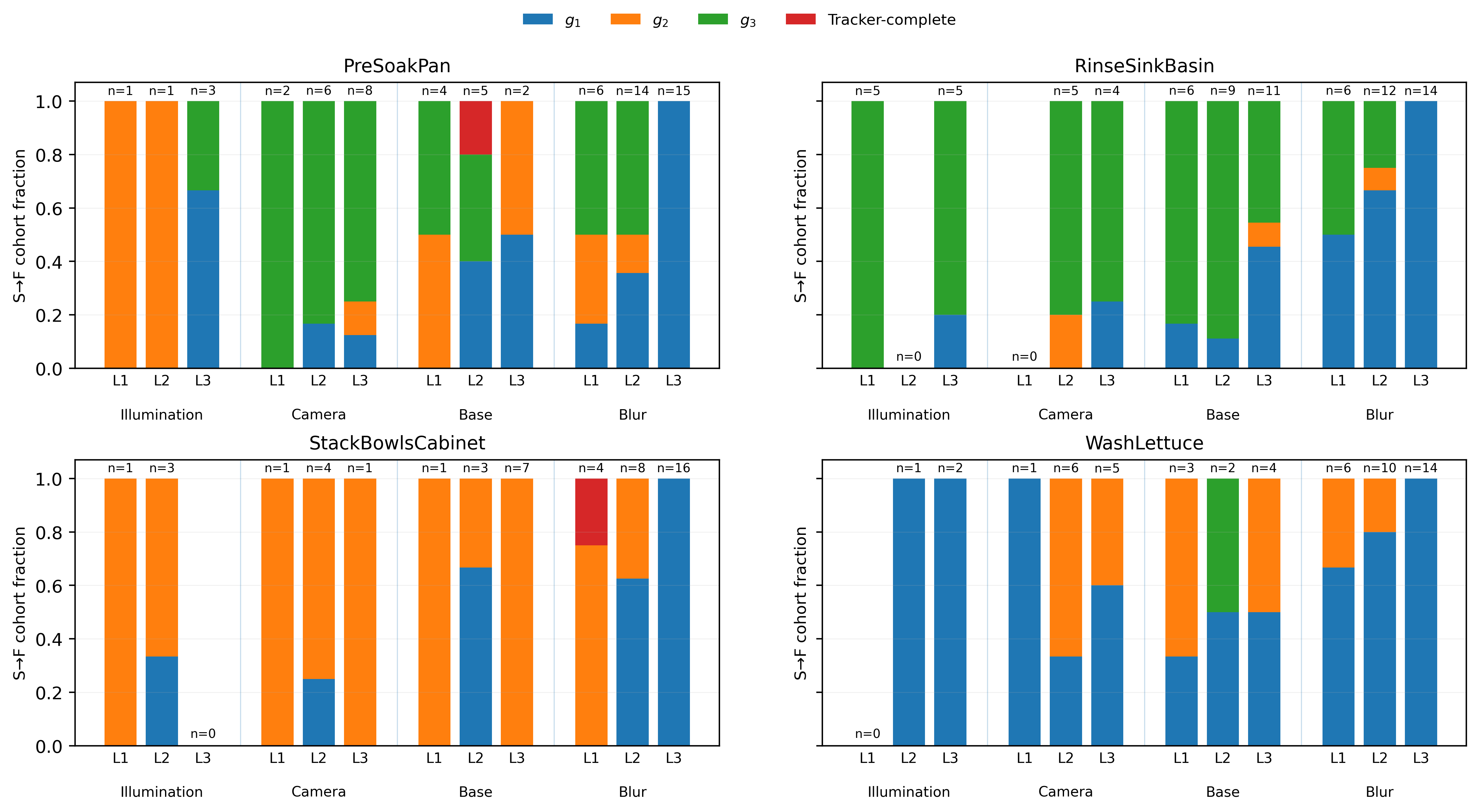}
    \caption{\textbf{Task-wise distributions of first incomplete
    semantic milestones among perturbation-induced failures.}
    Each subplot corresponds to one formal task. For each perturbation
    factor and severity level, the stacked bar shows the distribution
    of first incomplete milestones $g_1,\ldots,g_K$ among strictly
    matched $\mathrm{S}\!\rightarrow\!\mathrm{F}$ transitions,
    where $K$ depends on the task.
    Tracker-complete official failures, in which all tracked
    milestones are reached but RoboCasa's official task-success
    criterion remains unsatisfied, are reported separately
    without assigning a first incomplete milestone.
    The proportions are calculated over all
    $\mathrm{S}\!\rightarrow\!\mathrm{F}$ transitions in the
    corresponding task and condition.
    The annotation above each bar indicates the number of
    transitions $n$; conditions without such transitions
    are marked as $n=0$.}
    \label{fig:rq2_first_incomplete_appendix}
\end{figure*}

Table~\ref{tab:rq2_full_statistics} provides the complete
condition-level statistics for strictly matched
clean-success-to-perturbed-failure
($\mathrm{S}\!\rightarrow\!\mathrm{F}$) transitions,
including transition counts, mean normalized subgoal progress,
and first-stage failure fractions.
Figure~\ref{fig:rq2_first_incomplete_appendix} further
decomposes these transitions by task and first incomplete
semantic milestone, providing a task-specific view of the
stage-aware results presented in
Section~\ref{sec:results_failure_localization}.

\begin{table}[!b]
    \centering
    \caption{\textbf{Condition-level statistics for perturbation-induced
    failures.} Statistics are computed over strictly matched
    clean-success-to-perturbed-failure
    ($\mathrm{S}\!\rightarrow\!\mathrm{F}$) transitions.
    $N_{\mathrm{S}\rightarrow\mathrm{F}}$ denotes the number of
    transition events. First-stage reports the percentage of
    these transitions whose first incomplete tracked milestone
    is $g_1$.}
    \label{tab:rq2_full_statistics}
    \small
    \setlength{\tabcolsep}{4pt}
    \resizebox{\columnwidth}{!}{
    \begin{tabular}{llccc}
        \toprule
        Factor &
        Level &
        $N_{\mathrm{S}\rightarrow\mathrm{F}}$ &
        Mean Progress &
        First-stage (\%) \\
        \midrule

        Illumination
        & $L_1$ & 7  & 0.595 & 0.0 \\
        & $L_2$ & 5  & 0.267 & 40.0 \\
        & $L_3$ & 10 & 0.333 & 50.0 \\

        \midrule

        Camera Orientation
        & $L_1$ & 4  & 0.458 & 25.0 \\
        & $L_2$ & 21 & 0.437 & 19.0 \\
        & $L_3$ & 18 & 0.417 & 27.8 \\

        \midrule

        Base Initialization
        & $L_1$ & 14 & 0.464 & 14.3 \\
        & $L_2$ & 19 & 0.465 & 31.6 \\
        & $L_3$ & 24 & 0.340 & 33.3 \\

        \midrule

        Gaussian Blur
        & $L_1$ & 22 & 0.356 & 36.4 \\
        & $L_2$ & 44 & 0.223 & 59.1 \\
        & $L_3$ & 59 & 0.000 & 100.0 \\

        \bottomrule
    \end{tabular}}
\end{table}

The task-wise distributions reveal differences in the semantic
milestones left incomplete under the same perturbation condition.
For instance, under camera orientation at $L_2$, failures in
PreSoakPan and RinseSinkBasin predominantly leave $g_3$
incomplete, whereas those in WashLettuce are distributed
between $g_1$ and $g_2$.
Robot base initialization also exhibits task-dependent
distributions, including tracker-complete official failures
in PreSoakPan.
These observations complement the aggregate execution-depth
statistics by showing how perturbation-induced failures are
distributed across task-specific semantic milestones.

\subsection{Cross-Model Clean-Competence Screening}
\label{app:cross_model_screening}

We report the task-level clean-screening results for the additional
VLA policies considered in the cross-model extension.
Both SmolVLA RoboCasa and $\pi_0$ RoboCasa365 Human300 passed the
interface compatibility and fixed-seed repeatability checks before
clean screening. Table~\ref{tab:cross_model_clean_screen} compares
their screening outcomes on the frozen four-task suite with the
clean-qualified GR00T N1.5 reference.

\begin{table}[b]
    \centering
    \caption{\textbf{Clean-competence screening on the frozen
    four-task suite.}
    GR00T N1.5 results are from the second-round final clean
    screening ($N=20$ per task), whereas SmolVLA and $\pi_0$
    results are from the third-round initial screen
    ($N=10$ per task). Neither additional policy reached
    the clean-screen extension threshold on any task.}
    \label{tab:cross_model_clean_screen}
    \small
    \setlength{\tabcolsep}{5pt}
    \resizebox{\columnwidth}{!}{
    \begin{tabular}{lccc}
        \toprule
        Task &
        GR00T N1.5 &
        SmolVLA &
        $\pi_0$ \\
        \midrule

        PreSoakPan
            & 15/20 & 0/10 & 0/10 \\
        RinseSinkBasin
            & 14/20 & 0/10 & 0/10 \\
        StackBowlsCabinet
            & 16/20 & 0/10 & 2/10 \\
        WashLettuce
            & 14/20 & 1/10 & 0/10 \\

        \midrule
        Eligible tasks
            & 4/4 & 0/4 & 0/4 \\
        \bottomrule
    \end{tabular}}
\end{table}

To examine whether broader task coverage could support additional
evaluation, we also screened SmolVLA on a separately frozen
24-task pool comprising 12 mobile and 12 non-mobile tasks.
This screening contained 240 effective clean episodes,
including 20 records reused from the initial clean screening
on the frozen four-task evaluation suite.
Only OpenStandMixerHead and WashLettuce achieved one success
each out of ten episodes, while the remaining 22 tasks
recorded no successes. Consequently, none of the 24 tasks
qualified for clean-screen extension or subsequent
perturbation evaluation.
Two other released policies were considered but did not
enter clean screening due to interface compatibility constraints.
The official GR00T N1.7 base checkpoint did not support the
required \texttt{ROBOCASA\_PANDA\_OMRON} embodiment at the
policy-interface gate. Cosmos Policy, meanwhile, uses an
arm-only action interface and its released RoboCasa task
suite has no overlap with the four tasks used in our formal evaluation, preventing
evaluation under the existing protocol without additional,
unvalidated interface adaptations.
Under the frozen evaluation protocol, none of the additional
candidate policies progressed to perturbation evaluation.
These screening outcomes establish the limits of the available
policy-task coverage and should not be interpreted as a
cross-model robustness comparison or performance ranking.

\subsection{Additional Qualitative Counterfactual Examples}
\label{app:qualitative_counterfactual}

\begin{figure*}[t]
    \centering
    \includegraphics[
        width=\textwidth,
        height=0.78\textheight,
        keepaspectratio
    ]{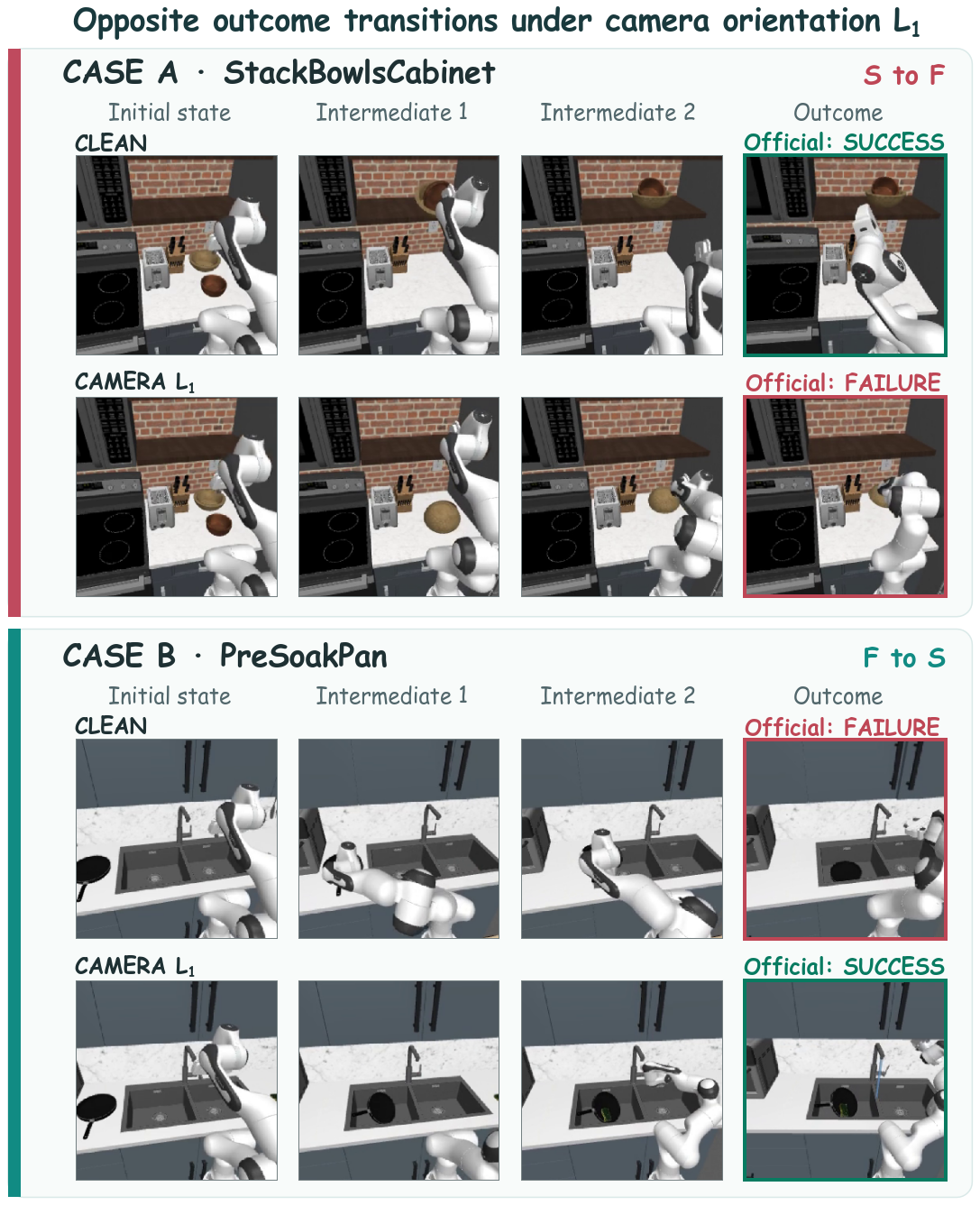}
    \caption{\textbf{Representative counterfactual outcome transitions
    under camera orientation at $L_1$.}
    Case A (StackBowlsCabinet) shows a success-to-failure
    ($\mathrm{S}\!\rightarrow\!\mathrm{F}$) transition, whereas
    Case B (PreSoakPan) shows a failure-to-success
    ($\mathrm{F}\!\rightarrow\!\mathrm{S}$) transition.
    Within each case, the clean and perturbed rollouts share
    the same task instance, initial simulator state, and
    episode-level policy seed, differing only in the target
    camera-orientation perturbation.
    Columns show selected frames from the initial, intermediate,
    and final stages of execution, with the final outcomes
    determined by RoboCasa's official task-success criterion.
    These examples illustrate opposite episode-level outcome
    changes under the same perturbation condition, despite
    zero aggregate success-rate degradation at $L_1$.}
    \label{fig:qualitative_opposite_transitions}
\end{figure*}

Figure~\ref{fig:qualitative_opposite_transitions} presents two
representative matched rollout pairs under camera orientation
at $L_1$, illustrating opposite directions of episode-level
outcome changes. While StackBowlsCabinet exhibits a
clean-success-to-perturbed-failure transition, PreSoakPan
shows the reverse transition. These examples complement
the paired transition statistics in
Section~\ref{sec:counterfactual_sensitivity} by providing
a qualitative view of outcome changes that can offset
each other in aggregate success-rate measurements.

\end{document}